\documentclass[letterpaper, 10pt,conference]{IEEEtran}

\usepackage{multicol}
\usepackage[bookmarks=true]{hyperref}
\hypersetup{
  linkcolor=blue
}

\usepackage{CJKutf8}
\usepackage{color}
\usepackage{enumerate}
\usepackage[ruled, linesnumbered]{algorithm2e}
\usepackage{amssymb, amsmath,mathtools}
\usepackage{stmaryrd}
\usepackage{graphicx}
\usepackage{setspace}
\usepackage{mathrsfs}
\usepackage{bm}
\usepackage[english]{babel}
\usepackage{blindtext}
\usepackage{epstopdf}
\usepackage[affil-it]{authblk}
\usepackage{cite}

\def\ie{i.e.\ }
\def\eg{e.g.\ }

\def\etc{etc.}

\usepackage{authblk}

\usepackage[top=0.75in,bottom=0.75in,right=0.75in,left=0.75in]{geometry}

\begin{document}

\title{Development of a Robotic System for\\Automated Decaking
  of 3D-Printed Parts}

\author[1,2*]{Huy Nguyen}
\author[1,2]{Nicholas Adrian}
\author[1,2]{Joyce Lim Xin Yan}
\author[1,3]{Jonathan M. Salfity}
\author[1,3]{William Allen}
\author[1,2]{Quang-Cuong Pham}
\affil[1]{HP-NTU Digital Manufacturing Corporate
  Lab, Nanyang Technological University,
  Singapore.}
\affil[2]{School of Mechanical and Aerospace Engineering, Nanyang
  Technological University, Singapore.}
\affil[3]{HP Labs, HP Inc.}
\affil[*]{Corresponding author. Email: huy.nguyendinh09@gmail.com}

\maketitle

\begin{abstract}
  With the rapid rise of 3D-printing as a competitive mass
  manufacturing method, manual ``decaking'' -- i.e. removing the
  residual powder that sticks to a 3D-printed part -- has become a
  significant bottleneck. Here, we introduce, for the first time to
  our knowledge, a robotic system for automated decaking of 3D-printed
  parts. Combining Deep Learning for 3D perception, smart mechanical
  design, motion planning, and force control for industrial robots, we
  developed a system that can automatically decake parts in a fast and
  efficient way. Through a series of decaking experiments performed on
  parts printed by a Multi Jet Fusion printer, we demonstrated the
  feasibility of robotic decaking for 3D-printing-based mass
  manufacturing.
\end{abstract}

\begin{IEEEkeywords}
deep learning, manipulation, system design, 3D-printing, decaking
\end{IEEEkeywords}

\section{Introduction}
\label{sec:introduction}

With the rapid rise of 3D-printing as a competitive mass manufacturing
method, the automated processing of 3D-printed parts has become a
critical and urgent need. Post-processing includes handling,
polishing, painting, assembling the parts that have been previously
3D-printed. In 3D-printing processes that involve a powder bed, such as
binder jet~\cite{sachs1993three,bai2015exploration}, Selective Laser
Sintering, Selective Laser Melting~\cite{olakanmi2015review} or HP
Multi Jet Fusion (MJF)~\cite{MJFwebsite}, a crucial post-processing
step is \emph{decaking}. After printing, some
non-bound/non-melted/non-sintered/non-fused powder usually sticks to
the 3D-printed part, forming together a ``cake''. Decaking consists of
removing that residual powder from the part, which can then be fed to
downstream post-processing stages.

\begin{figure}[t]
  \centering
  \includegraphics[width=0.35\textwidth]{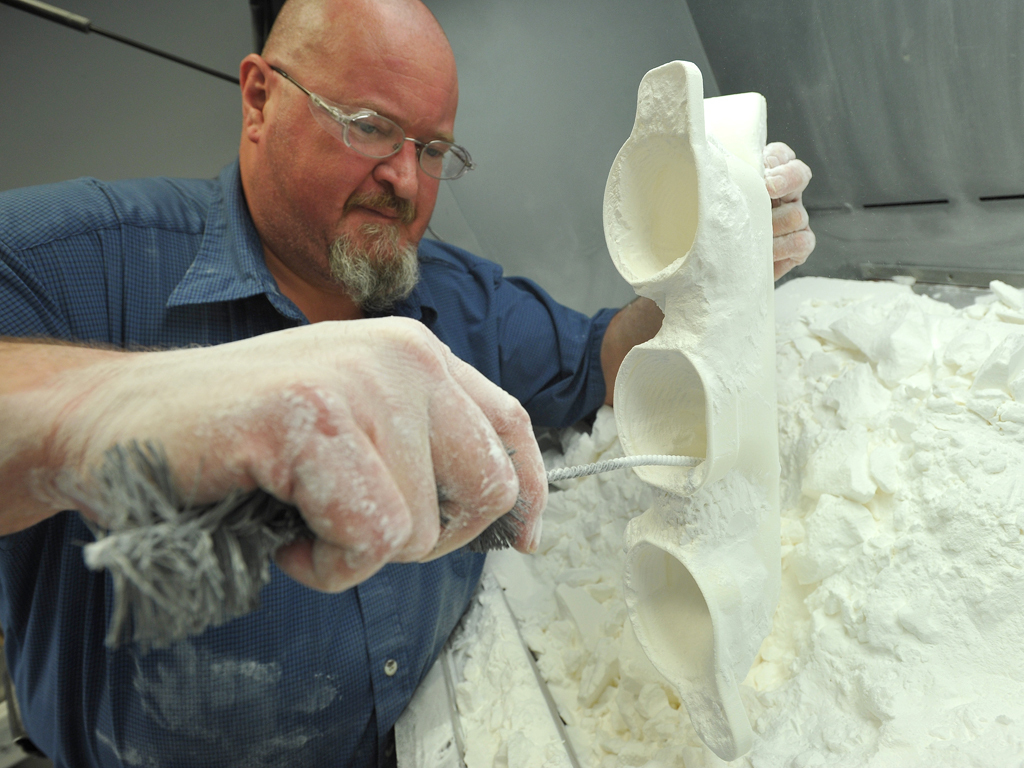}\\
  \vspace{0.1cm}
  \includegraphics[width=0.35\textwidth]{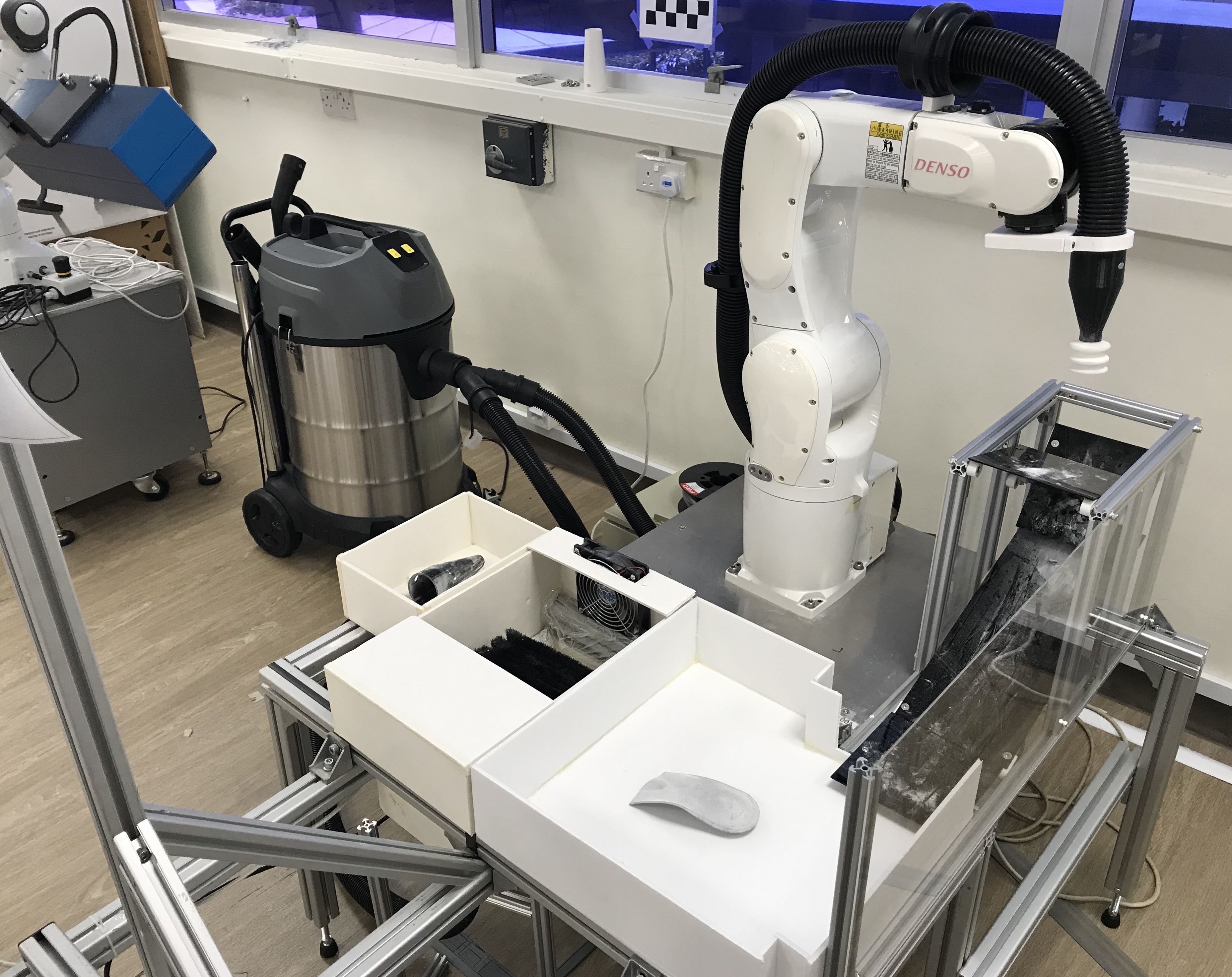}
  \caption{Top: An operator removing powder from a 3D-printed part
    (manual decaking). Bottom: Our proposed robotic system for
    automated decaking. A video of the actual robotic decaking process
    is available at \texttt{\url{https://youtu.be/0QJvNcf2s6s}}.}
  \label{fig:decaking}
\end{figure}

Decaking is currently done mostly by hand: human operators remove the
residual powder manually with brushes, as shown in
Figure~\ref{fig:decaking}, top. Recent 3D-printing technologies, such
as HP MJF, enable printing hundreds, or even thousands, of parts in a
single batch, making manual decaking a major bottleneck in
3D-printing-based mass manufacturing processes. There are some
automated approaches to decaking, using for instance
tumblers~\cite{ju2019improving}, but such approaches are unpractical
if the powder is moderately sticky or if the parts are fragile.

Here, we introduce, for the first time to our knowledge, a robotic
system for automated decaking of 3D-printed parts, see
Figure~\ref{fig:decaking}, bottom. Specifically, our system performs the
following steps:
\begin{enumerate}
\item Pick a caked part from the origin container with a suction cup;
\item Clean the underside of the caked part by rubbing it on a brush;
\item Flip the caked part;
\item Clean the other side of the caked part;
\item Place the cleaned part into the destination container.
\end{enumerate}

Note that those steps are very general and can be applied to other
3D-printing processes and to most parts that are nearly flat.

Step 1 is essentially a bin-picking task, which has recently drawn
significant attention from the community. However, the task at hand
presents unique difficulties: (i) the caked parts contain
unpredictable amount of residual powder, overlap each other, and are
mostly white-colored (the color of the powder), making part detection
and localization particularly challenging; (ii) the residual powder
and the parts have different physical properties, making caked part
manipulation, especially by a position-controlled industrial
robot\,\footnote{Most robots currently used in the industry are
  position-controlled, that is, they achieve highly accurate control
  in position and velocity, at the expense of poor, or no, control in
  force and torque. Yet, force or compliant control is crucial for
  contact tasks, such as part manipulation and decaking. A number of
  compliant robots have been developed in recent years but, compared
  to existing industrial robots, they are still significantly more
  expensive, less robust and more difficult to maintain.},
difficult. We address these challenges by leveraging respectively
(i)~recent advances in Deep Learning for 2D/3D vision; and (ii)~smart
mechanical design and force control.

Steps 2, 3 and 4 are specific to the decaking problem at hand. A
particular challenge is the control of the contacts between the
industrial robot, the parts, and the brushing system. We address this
challenge by emphasizing smart mechanical design whenever possible
(cleaning station, flipping station), and by using force
control~\cite{zeng1997overview} to perform compliant actions when
necessary. Our pipeline can be applied to all parts with nearly flat
shapes, such as shoe insoles\footnote{The authors have been granted
  permission by Footwork Podiatry Laboratory to use their shoe insole
  design and images in this paper.}, which, to date, are popular in
3D-printing-based mass manufacturing processes.

The remainder of the paper is organized as follows. In
Section~\ref{sec:relatedwork}, we review related work in robotic
cleaning, bin-picking, 3D perception, and force control. In
Section~\ref{sec:hardware}, we present our hardware system design.  In
Section~\ref{sec:software}, we describe our software stack. In
Section~\ref{sec:performance}, we evaluate the system performance in
number of actual decaking experiments, and demonstrate the feasibility
of automated decaking. Finally, in Section~\ref{sec:conclusion}, we
conclude and sketch some directions for future research.

\section{Related work}
\label{sec:relatedwork}
Recent years have seen a growing interest in
robotic surface/part cleaning. Advances in perception, control, planning and
learning have been applied to the problem~\cite{sato2011experimental,
  nagata2007cad, hess2012null, eppner2009imitation}. Most research
used robot manipulators to clean particles, powder or erase marker
ink~\cite{sato2011experimental, hess2012null, hess2014probabilistic, eppner2009imitation}, some considered mobile
robots to vacuum dust, or objects lying on the floor, such as papers
and clips~\cite{lawitzky2000navigation, jones2006robots}. In fact, factors like
variation in type of dust, shape of object, material, and
environmental constraints require different approaches and techniques
to tackle the problem. In this work, we are interested in the problem of cleaning 3D-printed parts which are randomly placed in a bin, and each part is covered by an unknown amount of residual or loose powder. Hence, we confine our review of prior work as follows:

As mentioned previously, the first step of our task is essentially a
bin-picking task, which is one of the classical problems in robotics
and have been investigated by many research groups,
\eg~\cite{drost2010model, buchholz2014combining, pretto2013flexible}. In these works, simplified
conditions are often exploited, \eg~parts with simple geometric primitives,
parts with holes to ease grasping action, parts with discriminative
features.

Since the Amazon Picking Challenge 2015, a variety of
approaches to a more general bin-picking problem had been proposed
\cite{eppner2016lessons,
  hernandez2016team, yu2016summary}. In~\cite{correll2016lessons},
authors discussed their observations and lessons drawn from a survey conducted among
participating teams in the challenge. Based on these lessons, we
identified 3D perception, which recognizes objects and determining their 3D poses in
a working space, as a key component to build a robust bin-picking system. Historically,
most approaches to conduct instance object recognition are based on local
descriptors around points of interest in images~\cite{lowe1999object}. While these
features are invariant to rotation, scale, and partial occlusions,
they are not very robust to illumination changes or partial
deformations of the object. Furthermore, since they depend on interest
points and local features, they only work well for objects with
texture. Recent advances in machine learning using convolutional neural
networks (CNNs) have rapidly improved object detection and segmentation
results~\cite{ren2015faster, he2017mask, liu2016ssd,
  redmon2017yolo9000}. Our detection module is based on a Mask R-CNN
model, which was originally open-sourced by Matterport Inc., and
implemented in TensorFlow and Keras~\cite{he2017mask}. The Mask R-CNN
model is an extension of the Faster R-CNN~\cite{ren2015faster} model
that achieved rapid object classification and bounding-box
regression. Mask R-CNN adopts its two-stage procedure from Faster
R-CNN. The first stage, called Regional Proposal Network (RPN),
proposes candidate regions in which there might be an object in the input
image. The second stage, in parallel, performs classification, refines
the bounding-box and generates a segmentation mask of the object on
each proposal region. Both stages are connected to a backbone
structure.

Another important conclusion is that system integration and development
are fundamental challenges to build task-specific, robust integrated
systems. It is desirable to study integrated solutions which include
all component technologies, such as 3D object pose estimation, control, motion
planning, grasping, etc. Besides, it is also highlighted that good
mechanical designs can sidestep challenging problems in control,
motion planning and object manipulation. In line with this idea, our
approach combines 3D perception using Deep Learning, smart mechanical
design, motion planning, and force control for industrial robots, to
develop a system that automatically decakes parts in a fast and
efficient way.


In addition, another challenging problem that is unique to our
decaking task is force
control. To achieve compliant motion control, we take a
position-controlled manipulator as a baseline system and make
necessary modifications by following the pipeline
in~\cite{suarez2018can, ott2010unified} (thanks to its robustness). 
This enables safe and sufficient cleaning operations without the
risk of damaging the parts. For more specific surveys on force control,
interested readers can refer to \cite{lefebvre2005active, calanca2015review}.

\section{Hardware system overview}
We design our system while keeping in mind a scalable solution that could eventually tackle the problem of
post-processing 3D-printed parts in a real scenario, at an advantageous cost. 
The robotic platform used in this work is, therefore,
characterized by cost-efficient, off-the-shelf components combined
with a classical position-control industrial
manipulator. In particular, the main components of our platform are

\begin{itemize}
\item 1 Denso VS060: Six-axis industrial manipulator;
\item 1 ATI Gamma Force-Torque (F/T) sensor; 
\item 1 Ensenso 3D camera N35-802-16-BL;
\item 1 suction system powered by a Karcher NT 70/2 vacuum machine;
\item 1 cleaning station;
\item 1 flipping station.
\end{itemize}

All control and computations are done on a computer with an Intel
Xeon E5-2630v3, 64 GB RAM.

\label{sec:hardware}
\begin{figure}[t]
  \centering
\includegraphics[width=0.5\textwidth]{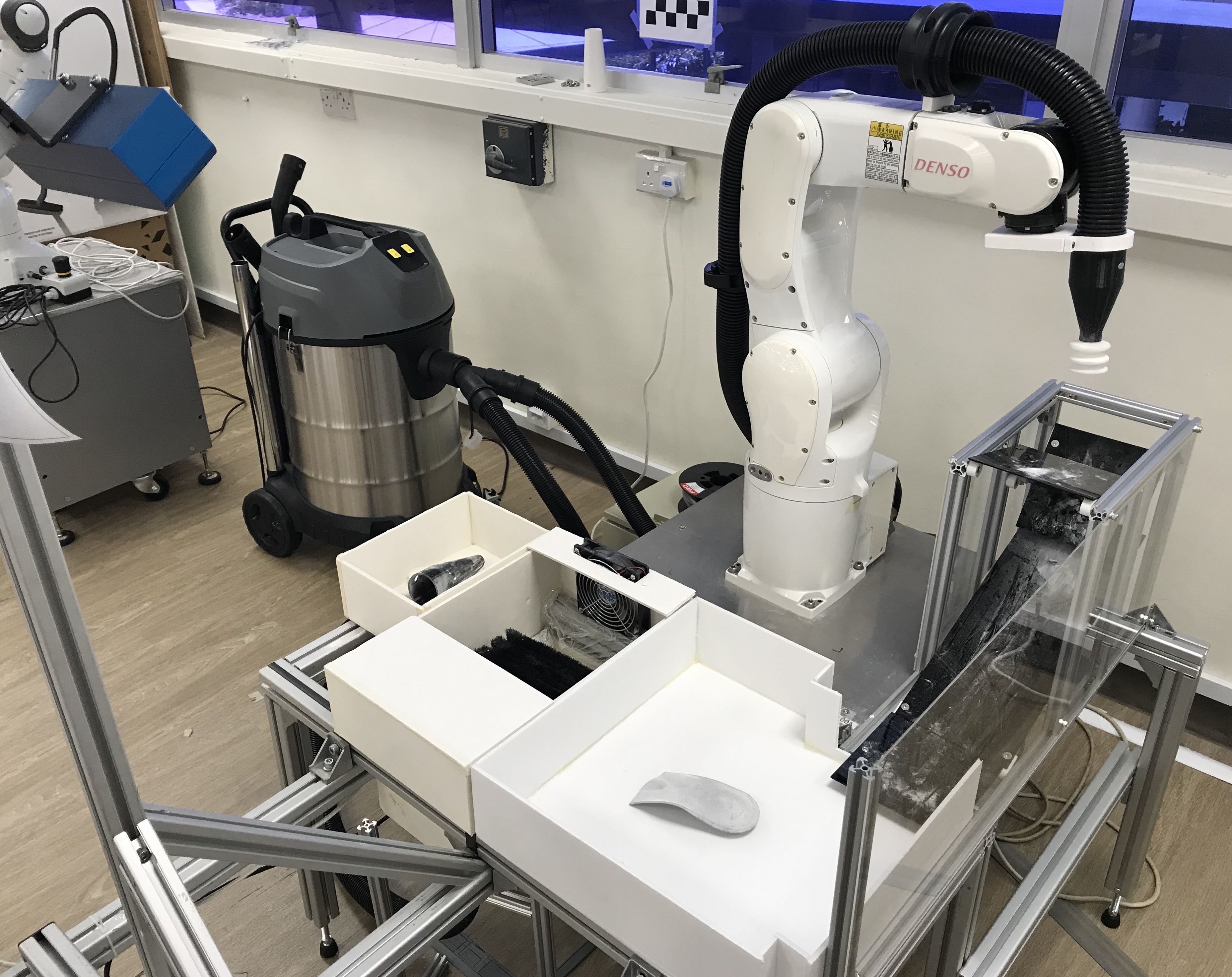}
  \caption{Our hardware system design}
  \label{fig:hardware}
\end{figure}


\subsection{Suction system}
\label{subsec:suction}
For grasping the parts, we decided to used a suction cup. This
choice was motivated by the versatility and performance of
suction-based system for bin-picking~\cite{correll2016lessons}, and by
fact that most
recovered powders are recyclable. Our suction system is designed to
generate both high vacuum and high air flow rate. This is to provide
sufficient force to lift parts and to maintain a firm hold of parts
during brushing. Air flow is guided from a suction cup on
the tip of the end effector through a flexible hose
into the vacuum machine. The vacuum itself is generated by a 2400W vacuum
cleaner. For binary on/off control, we
customize a 12V control output from the DENSO RC8 controller.

\subsection{Camera}
Robust perception is a one of the key components of the system. 
We manually optimized the selection of camera and camera location to
maximize the view angles; avoid occlusions from
robot arm, end-effector and cleaning, picking, dropping stations;
avoid collisions with objects and environment; and safety of the
device.

For the sake of robustness, our system employs Ensenso 3D camera
N35-802-16-BL and combines the use of both 2D grayscale image and
depth information.

\subsection{Cleaning station}
\begin{figure}[h]
  \centering
\includegraphics[width=0.4\textwidth]{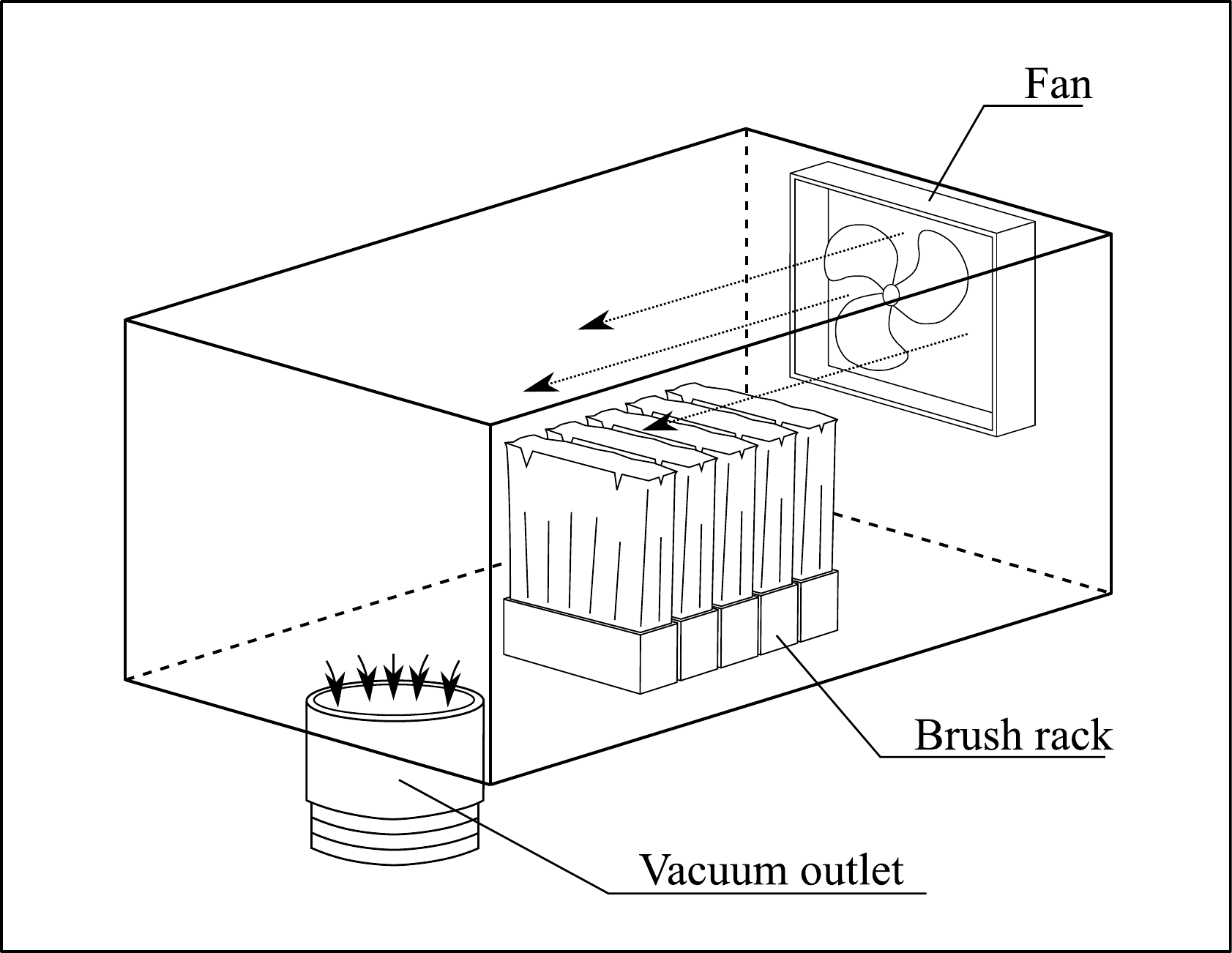}
  \caption{Cleaning station, comprising of a fan, a brush rack, and a
    vacuum outlet.}
  \label{fig:cleaning}
\end{figure}
To clean the parts, a brush rack is installed at the bottom of the cleaning station. We also integrate a dust
management system to collect excess powder removed during
brushing. Throughout the cleaning process, a blowing fan is
automatically activated to direct the powder stream frontward into the
vacuum suction hose, see Figure~\ref{fig:cleaning}.

\subsection{Flipping station}
\begin{figure}[h]
  \centering
\includegraphics[width=0.4\textwidth]{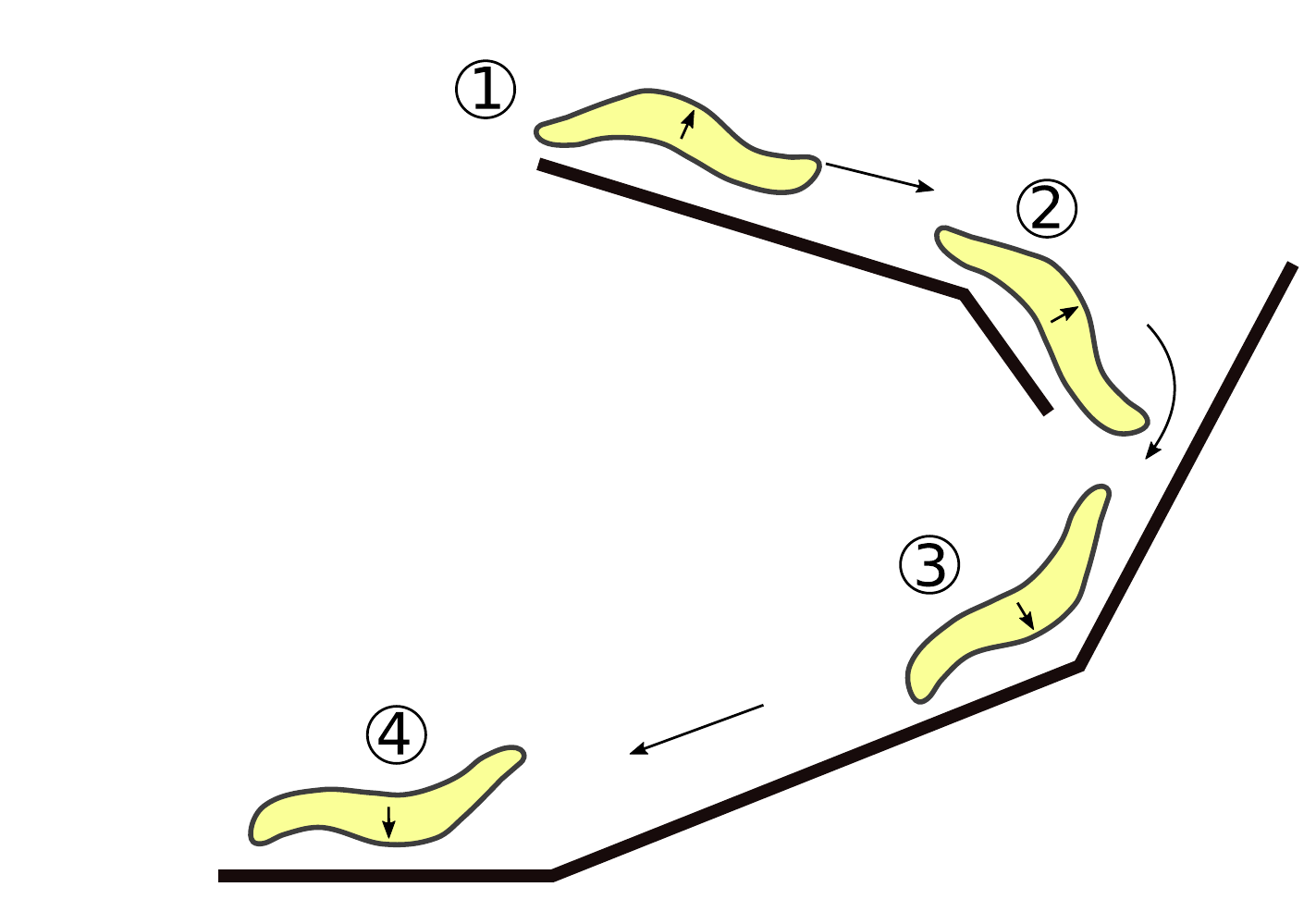}
  \caption{Flipping station.}
  \label{fig:flipping}
\end{figure}
To change the part orientation, we implement a passive flipping
station (\ie~no actuator is required to perform flipping). As illustrated
in Figure~\ref{fig:flipping}, the part is dropped from the top of the
station and
moves along the guiding sliders. Upon reaching the bottom of the
slider, the part is flipped
and is ready to be picked up by the robot.

This design is simple but works well with relatively flat parts such
as shoe insoles, flat boxes, etc. The next improvement of the station
will accommodate general part geometry and provide more reorientation
options.

\section{Software system overview}

\label{sec:software}
\begin{figure*}[t]
  \centering
  \includegraphics[width=0.85\textwidth]{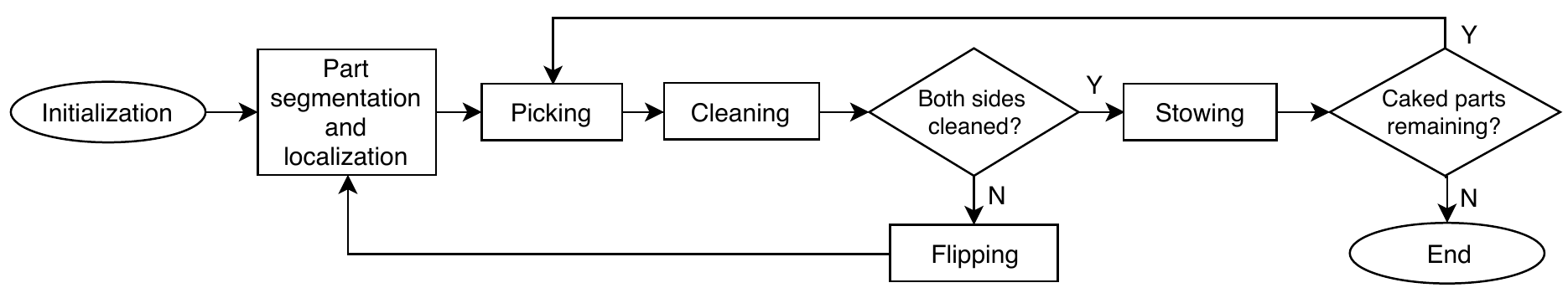}
  \caption{The system is composed of a state machine and a series of
    modules to perform perception and different types of action.}
  \label{fig:softwareschematic}
\end{figure*}

\begin{figure*}
  \centering
  \includegraphics[width=0.75\textwidth]{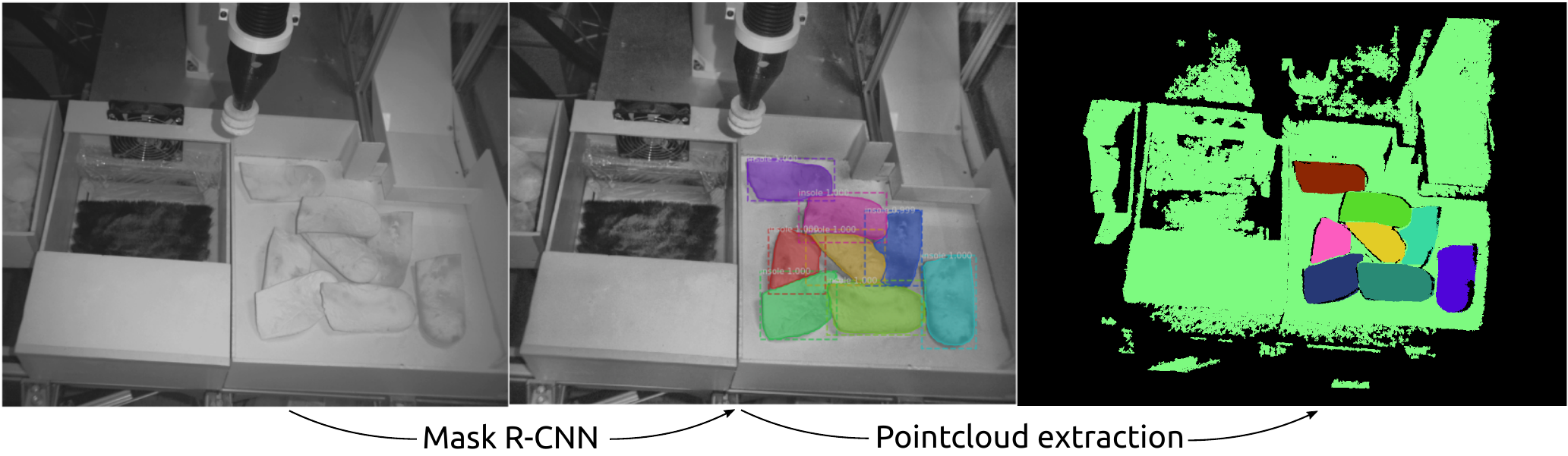}
  \caption{Example result of the object detection module based on Mask
    R-CNN. The estimated bounding boxes and part segmentations are
    depicted in different colors and labelled with the identification
    proposal and the confidence. We reject the detection with
    confidence lower than 95\%}
  \label{fig:detectionresult}
\end{figure*}

Our software system is composed of a state machine and a series of
modules. The communcation between different hardware and software
components is based on the Robot Operating System
(ROS)~\cite{quigley2009ros}.


\subsection{State machine}

The state machine is responsible for selecting the appropriate module
to execute at each point in time, see
Figure~\ref{fig:softwareschematic}. The solid arrows show the
direction to next states depending on the result of current states.
The state machine has access to all essential information of the
system, including types, poses, geometries and cleanliness, \etc~of
all objects detected in the scene. Each module can query this
information to realize its behavior. As a result, this design is
general and can be adapted to many more types of 3D-printed parts. For
example, the system can choose and execute a suitable cleaning
strategy based on the type of part, its pose, and its prioritized list
of actions; and decide whether to reattempt with a different strategy
or skip that part based on the outcome. In our implementation, the
modules were fine-tuned for each type of object based on previous
experiments to obtain a high success rate. It is desirable to have a
systematic approach to optimize the algorithm which decides the best
cleaning strategy to use for each type of part (see
Section~\ref{subsec:discussion}).

The state machine is also responsible for selecting the most feasible
part to be decaked next. Since the parts are often cluttered or
stacked, we generate the cleaning sequence by prioritizing parts with
large top surfaces and are on top of the clutter. This is to perform
suction easier and to de-stack parts in cluttered configurations.

\subsection{Perception}

The perception task is to identify and localize visible objects in the
working space. However, this is a challenging problem, due to heavy
occlusions (\ie~self-occlusion and occlusion by powder) and the poor
contrast between the freshly printed objects and their background. Our
approach is to separate the perception into two stages: object
detection, segmentation and 3D pose estimation. At the first stage, we
utilize a state-of-the-art deep-learning network to perform instance
detection and segmentation. The second stage extracts the 3D points of
each object using the segmentation mask and estimate the object pose.
This pipeline allow us to exploit the strength of deep-learning in
processing 2D images and depth information in estimating object
pose. We believe such combination of depth and RGB to be essential for
a robust perception solution.


First, a deep neural network based on Mask R-CNN classifies the
objects in the RGB image and performs instance segmentation, which
provides pixel-wise object classification. The developers had provided
a Mask R-CNN model that was pre-trained on large image classification
datasets (Microsoft COCO dataset~\cite{lin2014coco}). Hence, we
applied transfer learning on the pre-trained model to save training
time, by training the network to be able to classify a new object
class. In our experiment, this network requires from 75-100 labelled
images of real occurrences of typical 3D-printed parts. The result was
at very high detection rate of all parts presenting in the bin, as
shown in Figure~\ref{fig:detectionresult}. On average, our network has
a recall of 0.967 and a precision of 0.975. 

Second, pose estimation of the parts is done by estimating the
bounding boxes and computing the centroids of the segmented
pointclouds. The pointcloud of each object is refined (\ie statistical
outlier removal, normal smoothing, etc.) and used to verify if
the object can be picked by suction (\ie~exposed surfaces must be larger than suction
cup area). The 3D information of the bin is also used later during
the motion planning for collision avoidance. We included heuristics to
increase the robustness of the estimations by specifying physical
constraints inside the bin \eg~parts must lie within the bin's walls
and cannot be floating within the bin.

We also note that the Ensenso camera cannot capture both 2D and
3D images simultaneously as the high-intensity IR light tends to blind
the 2D camera. This module is, therefore, also in charge of turning on and off the
high-intensity IR strobe to prevent it from affecting 2D images.

\subsection{Motion primitives}

Some modules, such as ``Picking'' and ``Cleaning'', are themselves
composed of a series of motion primitives. 

\subsubsection*{Picking (suction-down) motion primitives}

These primitives are useful for picking parts with exposed nearly flat
surfaces. We limit our implementation to top-down directions. The
process is as follows:
\begin{enumerate}
\item The robot moves the suction cup above the centroid of the part
  and lowers the cup. Compliant force control is enabled and tells
  the robot when to stop the moving-down motion. 
\item Check whether the height at which the suction cup was stopped
  matches the expected height (from the point cloud with some
  tolerances).
\item The suction cup lifts, and the system constantly checks the
  force torque sensor to determine whether there is any collision
  (which often results large drag forces).
\end{enumerate}
Using suction is typically superior for parts with nearly flat and wide
surfaces. However, even when the suction cup could not form a perfect
seal with porous parts, high air flow often maintains sufficient force
on the part to lift it. Moreover, the suction cup is versatile and can
be replaced to fit many types of object.


\subsubsection*{Cleaning motion primitives}

These primitives remove residual powder and debris from the printed
parts. Our current implementation of these primitives is limited to
parts that are nearly flat (\eg~shoe insoles).

\begin{figure}
  \centering
  \includegraphics[width=0.4\textwidth]{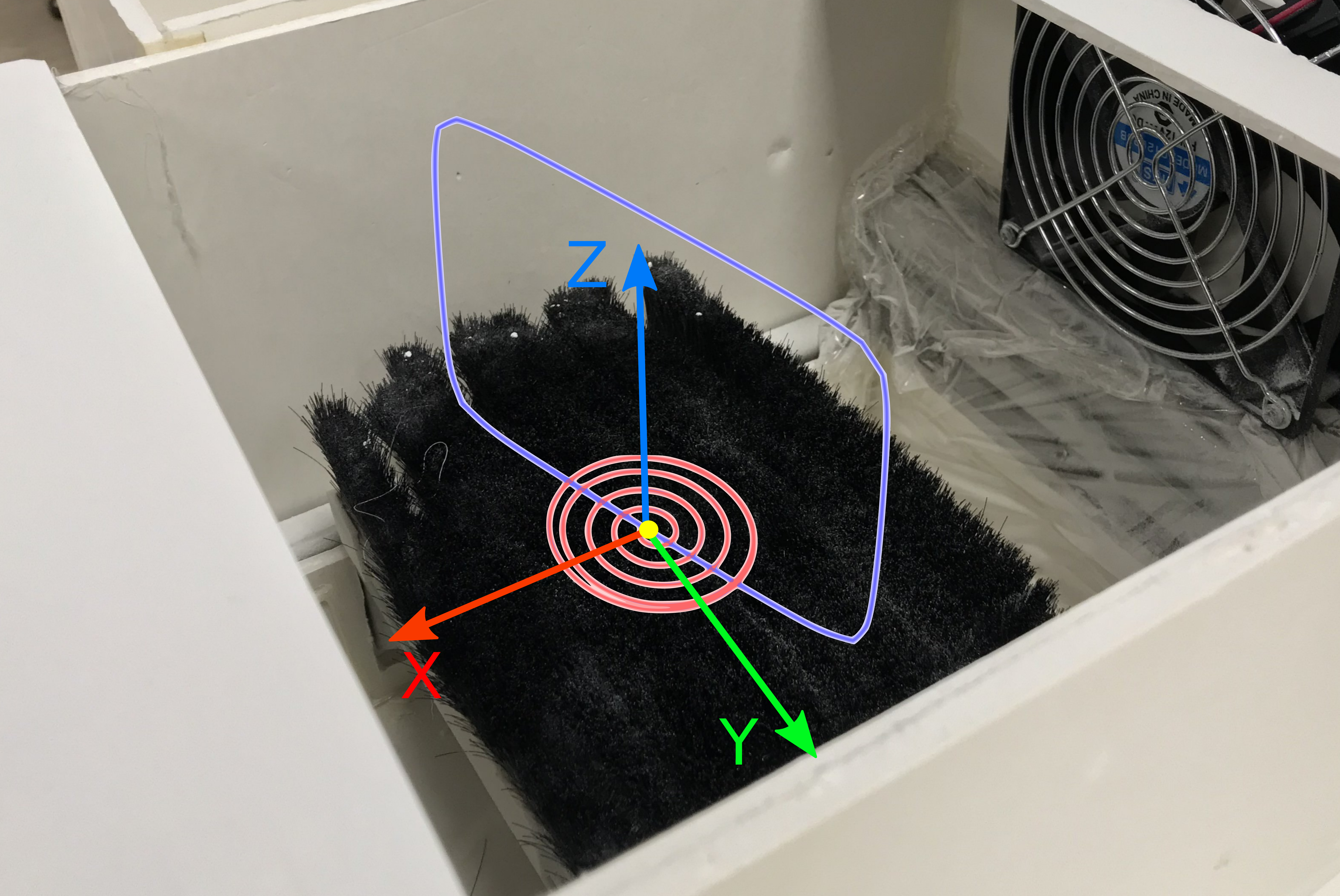}
  \caption{We use a combination of spiral and rectircle paths for
    cleaning motions. An example of spiral paths is shown in
    red. The yellow dot denotes the centroid of the parts at beginning
    of the motion. We modify the spiral paths so that they continue
    to circle around this yellow dot after reaching a maximum
    radius. An example of rectircle path is shown in blue. Rectircle
    path's parameters include width, height and its direction in XY
    plan.}
  \label{fig:cleaningpath}
\end{figure}

First, the robot positions the part above the brush rack in the
cleaning station. Then, the compliant force control mode is enabled to
move the robot until contact with the brushes. Next, the cleaning
trajectories are performed. To maintain constant contact between the
brushes and the part, we apply a hybrid position/force control scheme:
force is regulated in the direction normal to the brushes' surfaces,
while position is regulated in the tangential directions. The force
thresholds are determined through a number of trial-error
experiments. The cleaning trajectories are planned following two
patterns: \emph{spiral} and \emph{rectircle} (see
Figure~\ref{fig:cleaningpath}). While the spiral motion is well-suited
for cleanning nearly flat surfaces, the rectircle motion aids with
removing powder in concave areas.


\subsection{Motion planning and control}
All collision-free movements are planned online using the Bi-directional
Rapidly-Exploring Random Trees (RRTs) algorithm \cite{kuffner2000rrt} available in
OpenRAVE \cite{diankov2010thesis}.


\section{System performance and discussion}
\label{sec:performance} 
\subsection{Experimental setup}
\label{subsec:expsetup}
Parts considered in the experiments were 3D-printed shoe
insoles. At
the current stage of development, we assumed all parts to be unpacked from
printer powderbed and placed randomly in container bin. The
task of the system was to remove powder from the parts, which naturally consisted of several
key steps as follows:
\begin{enumerate}
\item Initial localization of parts in container bin,
\item Picking a part out of the container bin,
\item Removing residual powder from the underside of the part,
\item Flipping the part,
\item Re-localizing parts in the container bin and in the flipping collection area,
\item Picking flipped part,
\item Removing residual powder from the other side of the part,
\item Placing the part into the collection bin and continuing with any remaining
  parts (Step 2-8).
\end{enumerate}

\subsection{Performance}
\label{subsec:performance}

The sytem performed the cleaning task on 10 freshly 3D-printed shoe
insoles. Two aspects of the proposed system were experimentally
investigated. First, we evaluated the \emph{cleaning quality} by
weighing the parts before and after cleaning. Second, we reported the
actual \emph{running time} of the proposed system in a realistic
setting. As the baseline for comparison, we also
considered the same task performed by skilled human operators. The results of
this experiment are
summarized in Table~\ref{tab:performance}.

As expected, the performance of the system yielded a promising result
as a first prototype. Two test runs were conducted and successfully
executed by the proposed system. We noted that a perfectly cleaned
printed shoe insoles weighed $22.4\pm2$g on average. The proposed system
could remove powder from a dirty part on an average of $50.1\pm10.9$s and parts
weighed $37.6\pm6.4$g on average after cleaning, which meant $42.0\%\pm24.4\%$ excess
powder has been removed from the parts. In comparison, a skilled
human operator could clean $72.0\%\pm12.1\%$ of excess powder (parts weighed $29.8\pm3.2$g
after cleaning in $41.2\pm1.9$s on average). Although our system
performance regarding the cleaning quality was 1.7~times less than
that of a human operator, this was a potential outcome as our
robot system is supposed to work significantly longer. This however raised questions how task
efficiency could be further improved.

\subsection{Discussion}
\label{subsec:discussion}

We observed that our system performed actual brushing actions in only
$40\%$ of the execution time. In comparison, human spent more than
$95\%$ execution time on brushing. This was attributed to human
superior skills in performing sensing and dexterous
manipulations. However, when we limited brushing time to 20s, cleaning
quality was reduced as parts weighed $34.2\pm4.5$g on average after
cleaning (removed $55.5\%\pm17.9\%$ excess powder). This suggested
that one could improve cleaning quality by prolonging the brushing
duration and upgrading the cleaning station (\eg~spinning brushes to
speed up cleaning action; compliant brushes to clean concave areas,
etc.).

We also noted that humans provided more consistent results of cleaned
parts (\ie~parts cleaned by human operators had weight variation much
smaller than that of parts cleaned by our system). This was due to the
fact that human operators actively adjusted their cleaning motions
based on vision feedback. Hence, incorporating a cleanliness
evaluation module into the current system would improve the cleaning
quality performance. For this purpose, another 3D camera could be used
to observe the object after each cleaning section. The module would
retrieve a list of all dirty locations on object surfaces. This
information would allow the system to plan better cleaning motions and
could also be utilized to validate cleaning results.

Moreover, though this experiment featured a batch of nearly the same
3D-printed shoe insoles, our ultimate goal would be to extend this
robotic system to a various types of 3D-printed parts. For example, in
many cases, a batch of 3D-printed parts may contain a number of
different parts. The cleaning routines, hence, would have to be
altered to cater to the different geometries of each parts. To achieve
that, the state machine can be encoded with specific cleaning
instructions for each type of part. Much in the same way human
operators change cleaning routines for different parts, our robotic
system may recognize different parts, look up a prior defined handling
instructions, perform the cleanliness evaluation and execute
accordingly.

\begin{figure}
  \centering
  \includegraphics[width=0.47\textwidth]{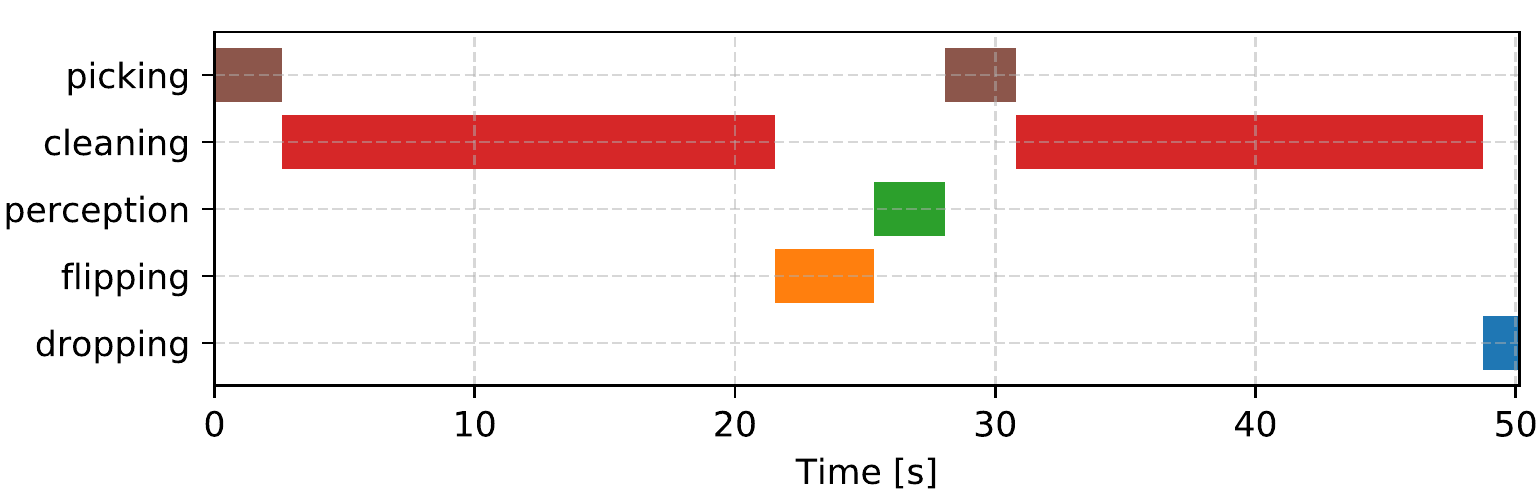}
  \caption{Average time-line representation of all actions used for the
    cleaning task.}
  \label{fig:timeline}
\end{figure}

To investigate the computational cost of our approach,
Figure~\ref{fig:timeline} also provides information on the average
times of each action used when performing the task. We noted that our
robot ran at $50\%$ max speed and all motions were planned
online. Hence, the sytem performance could be further enhanced by
optimizing these modules (\eg~off-line planning for tight-space
motions, working space optimization, etc.). Moreover, our perception
module was running on a CPU, implementations of better computing
hardware would thus improve the perception speed.

Furthermore, we are also working forward improving our end-effector
design. The current system is only capable of performing suction-down
motion primitives. Though it can accommodate for a very large tilting
angle, our system does not have enough maneuverability to perform side
picking. A possible approach would be to design a more generic suction
gripper to 
perform the task. Another solution would be to add another assisting
motion primitive, which is Toppling whose goal is not to pick, but to change the
configuration of an object that cannot be picked by the other
primitives. It is called in particular when the exposed face of an object
is too narrow. In this scenario it may be possible to topple the
object into a new configuration that will allow the object to be
picked by suction. Finally, air flow can be measured by using a pitot
tube inside the suction cup/hose. This is used to detect whether a
part was successfully grasped or lost during arm motions.

\begin{center}
  \begin{table}
    \small
  \caption{Performance of the proposed system, as compared to that of
    skilled human operators.}
  \centering
  \begin{tabular}{|p{2.35cm}|p{1.5cm}|p{1.6cm}|p{1.6cm}|}
    \hline
    Avg per part& Our system & Human
                               (no time-limit) & Human
                                                 (20s brushing)\\
    \hline
    Mass Before (g)&$48.6\pm10.9$&$48.8\pm7.8$&$48.9\pm8.0$\\
    \hline
    Mass After (g)&$37.6\pm6.4$&$29.8\pm3.2$&$34.2\pm4.5$\\
    \hline
    Cycle Time (s)&$50.1\pm2.1$&$41.2\pm1.9$&$21.2\pm1.1$\\
    \hline
    Brushing Time (s)&20&40&20\\
    \hline
  \end{tabular}
  \label{tab:performance}
\end{table}
\end{center}

\section{Conclusion}
\label{sec:conclusion}

In this paper, we have presented an automated robotic system for
decaking 3D-printed parts. We have combined Deep Learning for 3D
perception, smart mechanical design, motion planning and force control
for industrial robots to develop a system that can perform decaking
task in a fast and efficient manner. Through a series of experiments
performed on parts printed by a HP Multi Jet Fusion printer, we
demonstrated the feasibility of automatizing such tasks. The
achievements of this first prototype are promising and lay the
groundwork for multiple possible extensions: to other types of parts,
to other powder-based 3D-printing processes (SLM, SLS, binder jet,
etc.)  In future work, we will focus on validating the current system
further and expanding it to more general scenarios \eg~incorporating a
cleanliness evaluation module, optimizing task efficiency and adapting
the system to work on more general parts.
\section*{Acknowledgment}
This research was conducted in collaboration with HP Inc. and
partially supported by the Singapore Government through the Industry
Alignment Fund -Industry Collaboration Projects Grant. 
\pagebreak
\bibliographystyle{IEEEtran}
\IEEEtriggeratref{17}
\bibliography{robotic_decaking.bbl}

\begin{thebibliography}{10}
\providecommand{\url}[1]{#1}
\csname url@samestyle\endcsname
\providecommand{\newblock}{\relax}
\providecommand{\bibinfo}[2]{#2}
\providecommand{\BIBentrySTDinterwordspacing}{\spaceskip=0pt\relax}
\providecommand{\BIBentryALTinterwordstretchfactor}{4}
\providecommand{\BIBentryALTinterwordspacing}{\spaceskip=\fontdimen2\font plus
\BIBentryALTinterwordstretchfactor\fontdimen3\font minus
  \fontdimen4\font\relax}
\providecommand{\BIBforeignlanguage}[2]{{%
\expandafter\ifx\csname l@#1\endcsname\relax
\typeout{** WARNING: IEEEtran.bst: No hyphenation pattern has been}%
\typeout{** loaded for the language `#1'. Using the pattern for}%
\typeout{** the default language instead.}%
\else
\language=\csname l@#1\endcsname
\fi
#2}}
\providecommand{\BIBdecl}{\relax}
\BIBdecl

\bibitem{sachs1993three}
E.~M. Sachs, J.~S. Haggerty, M.~J. Cima, and P.~A. Williams,
  ``Three-dimensional printing techniques,'' Apr.~20 1993, {US} Patent
  5,204,055.

\bibitem{bai2015exploration}
Y.~Bai and C.~B. Williams, ``An exploration of binder jetting of copper,''
  \emph{Rapid Prototyping Journal}, vol.~21, no.~2, pp. 177--185, 2015.

\bibitem{olakanmi2015review}
E.~O. Olakanmi, R.~Cochrane, and K.~Dalgarno, ``A review on selective laser
  sintering/melting {(SLS/SLM)} of aluminium alloy powders: Processing,
  microstructure, and properties,'' \emph{Progress in Materials Science},
  vol.~74, pp. 401--477, 2015.

\bibitem{MJFwebsite}
\BIBentryALTinterwordspacing
hp.com. (2019) {HP} multi jet fusion technology. [Online]. Available:
  \url{https://www8.hp.com/sg/en/printers/3d-printers/products/multi-jet-technology.html}
\BIBentrySTDinterwordspacing

\bibitem{ju2019improving}
A.~Ju, A.~Fitzhugh, J.~Jun, and M.~Baker, ``Improving aesthetics through
  post-processing for {3D} printed parts,'' \emph{Electronic Imaging}, vol.
  2019, no.~6, pp. 480--1, 2019.

\bibitem{zeng1997overview}
G.~Zeng and A.~Hemami, ``An overview of robot force control,'' \emph{Robotica},
  vol.~15, no.~5, pp. 473--482, 1997.

\bibitem{sato2011experimental}
F.~Sato, T.~Nishii, J.~Takahashi, Y.~Yoshida, M.~Mitsuhashi, and D.~Nenchev,
  ``Experimental evaluation of a trajectory/force tracking controller for a
  humanoid robot cleaning a vertical surface,'' in \emph{2011 IEEE/RSJ
  international conference on intelligent robots and systems}, 2011, pp.
  3179--3184.

\bibitem{nagata2007cad}
F.~Nagata, T.~Hase, Z.~Haga, M.~Omoto, and K.~Watanabe, ``{CAD/CAM-based}
  position/force controller for a mold polishing robot,'' \emph{Mechatronics},
  vol.~17, no. 4-5, pp. 207--216, 2007.

\bibitem{hess2012null}
J.~Hess, G.~D. Tipaldi, and W.~Burgard, ``Null space optimization for effective
  coverage of 3d surfaces using redundant manipulators,'' in \emph{IEEE/RSJ
  International Conference on Intelligent Robots and Systems}, 2012, pp.
  1923--1928.

\bibitem{eppner2009imitation}
C.~Eppner, J.~Sturm, M.~Bennewitz, C.~Stachniss, and W.~Burgard, ``Imitation
  learning with generalized task descriptions,'' in \emph{International
  Conference on Robotics and Automation}.\hskip 1em plus 0.5em minus
  0.4em\relax IEEE, 2009, pp. 3968--3974.

\bibitem{hess2014probabilistic}
J.~Hess, M.~Beinhofer, and W.~Burgard, ``A probabilistic approach to
  high-confidence cleaning guarantees for low-cost cleaning robots,'' in
  \emph{International conference on robotics and automation (ICRA)}.\hskip 1em
  plus 0.5em minus 0.4em\relax IEEE, 2014, pp. 5600--5605.

\bibitem{lawitzky2000navigation}
G.~Lawitzky, ``A navigation system for cleaning robots,'' \emph{Autonomous
  Robots}, vol.~9, no.~3, pp. 255--260, 2000.

\bibitem{jones2006robots}
J.~L. Jones, ``Robots at the tipping point: the road to irobot roomba,''
  \emph{IEEE Robotics \& Automation Magazine}, vol.~13, no.~1, pp. 76--78,
  2006.

\bibitem{drost2010model}
B.~Drost, M.~Ulrich, N.~Navab, and S.~Ilic, ``Model globally, match locally:
  Efficient and robust 3d object recognition,'' in \emph{Computer society
  conference on computer vision and pattern recognition}.\hskip 1em plus 0.5em
  minus 0.4em\relax IEEE, 2010, pp. 998--1005.

\bibitem{buchholz2014combining}
D.~Buchholz, D.~Kubus, I.~Weidauer, A.~Scholz, and F.~M. Wahl, ``Combining
  visual and inertial features for efficient grasping and bin-picking,'' in
  \emph{International Conference on Robotics and Automation (ICRA)}.\hskip 1em
  plus 0.5em minus 0.4em\relax IEEE, 2014, pp. 875--882.

\bibitem{pretto2013flexible}
A.~Pretto, S.~Tonello, and E.~Menegatti, ``Flexible {3D} localization of planar
  objects for industrial bin-picking with monocamera vision system,'' in
  \emph{International Conference on Automation Science and Engineering
  (CASE)}.\hskip 1em plus 0.5em minus 0.4em\relax IEEE, 2013, pp. 168--175.

\bibitem{eppner2016lessons}
C.~Eppner, S.~H{\"o}fer, R.~Jonschkowski, R.~Mart{\'\i}n-Mart{\'\i}n,
  A.~Sieverling, V.~Wall, and O.~Brock, ``Lessons from the amazon picking
  challenge: Four aspects of building robotic systems.'' in \emph{Robotics:
  Science and Systems}, 2016.

\bibitem{hernandez2016team}
C.~Hernandez, M.~Bharatheesha \emph{et~al.}, ``Team delft’s robot winner of
  the amazon picking challenge 2016,'' in \emph{Robot World Cup}.\hskip 1em
  plus 0.5em minus 0.4em\relax Springer, 2016, pp. 613--624.

\bibitem{yu2016summary}
K.-T. Yu, N.~Fazeli, N.~Chavan-Dafle, O.~Taylor, E.~Donlon, G.~D. Lankenau, and
  A.~Rodriguez, ``A summary of team {MIT's} approach to the amazon picking
  challenge 2015,'' \emph{arXiv preprint arXiv:1604.03639}, 2016.

\bibitem{correll2016lessons}
N.~Correll, K.~E. Bekris, D.~Berenson, O.~Brock, A.~Causo, K.~Hauser, K.~Okada,
  A.~Rodriguez, J.~M. Romano, and P.~R. Wurman, ``Lessons from the amazon
  picking challenge,'' \emph{arXiv preprint arXiv:1601.05484}, vol.~3, 2016.

\bibitem{lowe1999object}
D.~Lowe, ``Object recognition from local scale-invariant features.'' in
  \emph{International Conference on Computer Vision}, vol.~99, no.~2.\hskip 1em
  plus 0.5em minus 0.4em\relax IEEE, 1999, pp. 1150--1157.

\bibitem{ren2015faster}
S.~Ren, K.~He, R.~Girshick, and J.~Sun, ``Faster {R-CNN}: Towards real-time
  object detection with region proposal networks,'' in \emph{Advances in neural
  information processing systems}, 2015, pp. 91--99.

\bibitem{he2017mask}
K.~He, G.~Gkioxari, P.~Doll{\'a}r, and R.~Girshick, ``Mask {R-CNN},'' in
  \emph{Proceedings of the IEEE international conference on computer vision},
  2017, pp. 2961--2969.

\bibitem{liu2016ssd}
W.~Liu, D.~Anguelov, D.~Erhan, C.~Szegedy, S.~Reed, C.-Y. Fu, and A.~C. Berg,
  ``{SSD}: Single shot multibox detector,'' in \emph{European conference on
  computer vision}.\hskip 1em plus 0.5em minus 0.4em\relax Springer, 2016, pp.
  21--37.

\bibitem{redmon2017yolo9000}
J.~Redmon and A.~Farhadi, ``{YOLO9000:} better, faster, stronger,'' in
  \emph{Proceedings of the IEEE conference on computer vision and pattern
  recognition}, 2017, pp. 7263--7271.

\bibitem{suarez2018can}
F.~Su{\'a}rez-Ruiz, X.~Zhou, and Q.-C. Pham, ``Can robots assemble an {IKEA}
  chair?'' \emph{Science Robotics}, vol.~3, no.~17, p. eaat6385, 2018.

\bibitem{ott2010unified}
C.~Ott, R.~Mukherjee, and Y.~Nakamura, ``Unified impedance and admittance
  control,'' in \emph{International Conference on Robotics and
  Automation}.\hskip 1em plus 0.5em minus 0.4em\relax IEEE, 2010, pp. 554--561.

\bibitem{lefebvre2005active}
T.~Lefebvre, J.~Xiao, H.~Bruyninckx, and G.~De~Gersem, ``Active compliant
  motion: a survey,'' \emph{Advanced Robotics}, vol.~19, no.~5, pp. 479--499,
  2005.

\bibitem{calanca2015review}
A.~Calanca, R.~Muradore, and P.~Fiorini, ``A review of algorithms for compliant
  control of stiff and fixed-compliance robots,'' \emph{IEEE/ASME Transactions
  on Mechatronics}, vol.~21, no.~2, pp. 613--624, 2015.

\bibitem{quigley2009ros}
M.~Quigley, K.~Conley, B.~Gerkey, J.~Faust, T.~Foote, J.~Leibs, R.~Wheeler, and
  A.~Y. Ng, ``{ROS:} an open-source robot operating system,'' in \emph{ICRA
  workshop on open source software}, vol.~3, no. 3.2.\hskip 1em plus 0.5em
  minus 0.4em\relax Kobe, Japan, 2009, p.~5.

\bibitem{lin2014coco}
T.-Y. Lin, M.~Maire, S.~Belongie, J.~Hays, P.~Perona, D.~Ramanan,
  P.~Doll{\'a}r, and C.~L. Zitnick, ``Microsoft {COCO}: Common objects in
  context,'' in \emph{European conference on computer vision}.\hskip 1em plus
  0.5em minus 0.4em\relax Springer, 2014, pp. 740--755.

\bibitem{kuffner2000rrt}
J.~J. Kuffner~Jr and S.~M. LaValle, ``{RRT-connect:} an efficient approach to
  single-query path planning,'' in \emph{International Conference on Robotics
  and Automation}, vol.~2.\hskip 1em plus 0.5em minus 0.4em\relax IEEE, 2000.

\bibitem{diankov2010thesis}
R.~Diankov, ``Automated construction of robotic manipulation programs,'' Ph.D.
  dissertation, Carnegie Mellon University, 2010.

\end{thebibliography}

\end{document}